# Towards Effective Human-AI Decision-Making: The Role of Human Learning in Appropriate Reliance on AI Advice

*Completed Research Paper*

**Max Schemmer, Andrea Bartos,**
**Philipp Spitzer, Patrick Hemmer,**
**Jonas Liebschner, Gerhard Satzger**
Karlsruhe Institute of Technology,
Karlsruhe, Germany
max.schemmer@kit.edu, andrea.bartos@alumni.kit.edu,
philipp.spitzer@kit.edu, patrick.hemmer@kit.edu,
jonas.liebschner@alumni.kit.edu, gerhard.satzger@kit.edu

**Niklas Kühl**
University of Bayreuth
Bayreuth, Germany
kuehl@uni-bayreuth.de

## Abstract

*The true potential of human-AI collaboration lies in exploiting the complementary capabilities of humans and AI to achieve a joint performance superior to that of the individual AI or human, i.e., to achieve complementary team performance (CTP). To realize this complementarity potential, humans need to exercise discretion in following AI's advice, i.e., appropriately relying on the AI's advice. While previous work has focused on building a mental model of the AI to assess AI recommendations, recent research has shown that the mental model alone cannot explain appropriate reliance. We hypothesize that, in addition to the mental model, human learning is a key mediator of appropriate reliance and, thus, CTP. In this study, we demonstrate the relationship between learning and appropriate reliance in an experiment with 100 participants. This work provides fundamental concepts for analyzing reliance and derives implications for the effective design of human-AI decision-making.*

**Keywords:** Human-AI Decision-Making, Human-AI Complementarity, Appropriate Reliance, Organizational Learning

## Introduction

Over the past years, Artificial Intelligence (AI) systems have entered a wide range of areas, even high-stake decision domains. For instance, AI applications support doctors in their diagnoses (Leibig et al. 2022), help recruiters in the hiring process (Peng et al. 2022), and support legal decisions in court (Kleinberg et al. 2018). This proliferation is driven by the continuous development of AI systems, which results in advanced capabilities and increased performance (Ren et al. 2015). Self-supervised models and generative AI have further fueled a new era of human-AI decision-making, e.g., Notion AI, GitHub Co-pilot, DeepL, and ChatGPT.

Modern AI is not only more performant and versatile in its applications but also bears the potential to enhance humans through complementary capabilities (Dellermann et al. 2019; Fügener et al. 2021; Hemmer et al. 2021), reaching performance levels beyond the one's humans or AI can reach on their own. This desired superior performance in human-AI decision-making is referred to as complementary team performance (CTP) (Bansal et al. 2021; Hemmer et al. 2021).

Despite the tremendous advances in performance and capabilities, it is essential to bear in mind that every AI application has inherent uncertainty. AI models and their recommendations are based on probabilities. So, no matter how good a model is, it will not always be accurate. Since AI advice is imperfect, general acceptance by a human decision-maker would also comprise incorrect advice. For example, physicians would blindly follow AI advice on cancer diagnosis—although AI advice might be wrong, and the physicians might have known better. Thus, it is important for human decision-makers to have the ability to discern when to rely on AI advice and when to rely on their judgment, i.e., they should display a high level of appropriateness *of reliance* (AoR). AoR is a reliance metric based on the relative frequency of correctly overriding incorrect AI advice (self-reliance) and following correct AI advice (AI-reliance) and is the link between reliance behavior and CTP (Schemmer et al. 2023).

So far, research has focused on driving AoR by enabling humans to build an accurate mental model of AI (Bansal et al. 2019; Kloker et al. 2022; Kuhl et al. 2020; Taudien et al. 2022)). The term mental model refers to the human's knowledge about various aspects of the AI system's capabilities. Bansal et al. (2019) particularly stress the importance of recognizing the AI's error boundaries to develop a realistic mental model. The mental model research stream focuses on enabling decision-makers to assess the quality of an AI prediction. Researchers aim to build this mental model by providing explanations of the AI's decision process (Bansal et al. 2021; Zhang et al. 2020).

However, recent review articles have shown that the current focus on influencing the mental model through the provisioning of explanations is not sufficient to consistently improve AoR and reach CTP (Bansal et al. 2021; Hemmer et al. 2021; Schemmer, Hemmer, Nitsche, et al. 2022). For example, in a behavioral experiment, Fügener et al. (2023) find that providing uncertainty information of AI does not lead to an appropriate level of human reliance on AI.

We hypothesize that the mental model is not the only mediator influencing AoR and that researchers need to investigate further impact factors. One of the core influence factors of reliance behavior in human-AI decision-making seems to be whether the decision-maker is an expert (high domain knowledge) or a lay worker (low domain knowledge) (Nourani et al. 2020; Wang and Yin 2021). We follow this line of thought and hypothesize that task-specific *learning* during human-AI decision-making (decision-makers gradually gaining expertise) could be a relevant mediator of AoR.

However, the specific effect of learning on AoR is ambiguous. Learning during collaboration could also lead to unwanted effects, such as aversion—as humans might think they have learned enough to solve the task alone and no longer need the AI advice. Therefore, we formulate the following research question.

> **RQ1:** How does human task-specific learning during human-AI decision-making influence the Appropriateness of Reliance on AI advice?

Even if learning improves AoR, we must find ways to enable and improve learning during human-AI decision-making. Recent research on learning systems has seen first positive results of using explanations of AI to improve learning (Goyal et al. 2019; Wang and Vasconcelos 2020). However, it is an open question if the promising result in learning systems can be replicated in human-AI decision-making. Therefore, we formulate our second research question.

> **RQ2:** How do explanations of AI influence human task-specific learning in human-AI decision-making?

To gain first answers to our broader research questions, we derive a research model including theory-driven hypotheses and subsequently conduct a behavioral experiment with 100 participants using an image classification task as a testbed to evaluate the model. We use example-based explanations (Fahse et al. 2022) to design a human-AI decision-making scenario with a high potential for learning.

Our results show that a) example-based explanations can improve human learning during human-AI decision-making, b) learning improves the human ability to assess when to rely on themselves, and c) if sufficient learning is present, human learning helps to assess better when to rely on AI.

We contribute to the body of knowledge on human-AI decision-making in general and on AoR and learning from AI in particular. To the best of our knowledge, this research depicts the first study covering the effect of explanations on learning and the mediating effect on AoR. We thereby extend the research model of AoR developed by Schemmer et al. (2023) by a learning construct. In addition, we contribute to the organizational learning research stream (Levitt and March 1988) by showing that it is possible to learn domain knowledge during human-AI decision-making. Our work provides a new perspective on AoR and the design of human-AI decision-making systems.

# Theoretical Foundations & Related Work

In the following, we introduce the related work of this article, structured along the topics of human-AI decision-making and learning from AI. In the first subsection, we introduce human-AI decision-making and explain AoR. Next, we provide foundations about task-specific learning from AI, our hypothesized mediator of AoR, and how to influence task-specific learning by providing explanations of AI.

## *Human-AI Decision-Making and Appropriate Reliance*

In recent years, there has been a surge of research in human-AI decision-making, with a growing number of studies conducting behavioral experiments to gain a better understanding of how humans form decisions in the presence of AI (Alufaisan et al. n.d.; Buçinca et al. 2020; Carton et al. 2020; Lai et al. 2020; Lai and Tan 2019; Liu et al. 2021; Zhang et al. 2020). This research has focused on improving human-AI decision-making to optimize team performance (Buçinca et al. 2020; Zhang et al. 2020).

The idea behind human-AI decision-making is to be more effective than both human and AI individually (Dellermann et al. 2019). This improvement through collaboration stems from the idea that both the AI and the human possess a unique set of skills that can enrich each other in specific tasks. Thus, the true potential of human-AI decision-making lies in leveraging these complementary capabilities to reach the desired state of superior performance, i.e., complementary team performance (CTP) (Bansal et al. 2021; Hemmer et al. 2021).

The question of how to realize this complementarity potential and thus reach the desired superior performance remains an active area of research. In most empirical studies, the performance of human-AI decision-making is still inferior to that of individual AI, and thus CTP is not achieved (Bansal et al. 2021; Hemmer et al. 2021; Schemmer, Hemmer, Nitsche, et al. 2022). Current research identifies the missing appropriateness of reliance as a main cause preventing the achievement of CTP (Bansal et al. 2021; Schemmer, Hemmer, Nitsche, et al. 2022).

The concept of appropriate reliance has gained attention in the field of human-AI decision-making (Bansal et al. 2021; Schemmer, Hemmer, Nitsche, et al. 2022). In general, appropriate reliance refers to desirable behavior where humans override incorrect AI advice and follow correct advice (Bansal et al. 2021; Schemmer et al. 2023).

First work on the conceptualization and measurement of appropriate reliance was done by Schemmer et al. (2023). The authors differentiate between appropriate reliance as a binary target state ("appropriate reliance is either achieved or not") and a metric indicating a degree of appropriateness. They introduce a two-dimensional metric—the appropriateness of reliance (AoR)[1]—to describe and measure reliance behavior. It is based on relative frequencies of correctly overriding wrong AI suggestions (correct self-reliance) and following correct AI suggestions (correct AI reliance) and reflects a metric understanding of appropriate reliance (see Figure 1 on page 4). This metric can then be used to define different levels as target states of appropriate reliance that mark the achievement of objectives like certain legal, ethical and performance requirements. From an effectiveness perspective they argue that appropriate reliance is achieved if CTP is present.

---

[1] Note that the measurement of AoR requires a sequential task setup as visualized in Figure 1 on page 4 and described in Schemmer et al. (2023).

Bansal et al. (2019) emphasize the importance of humans having an accurate understanding of what the AI system can and cannot do to enable appropriate reliance. This understanding, referred to as the "mental model", encompasses various facets of the AI's capabilities. If there's a disparity between a human's mental model and the actual error limitations of the AI, this can result in poor decisions—either misplaced reliance in the AI when it is incorrect, or unwarranted skepticism when it is actually correct (Taudien et al. 2022), i.e., low AoR.

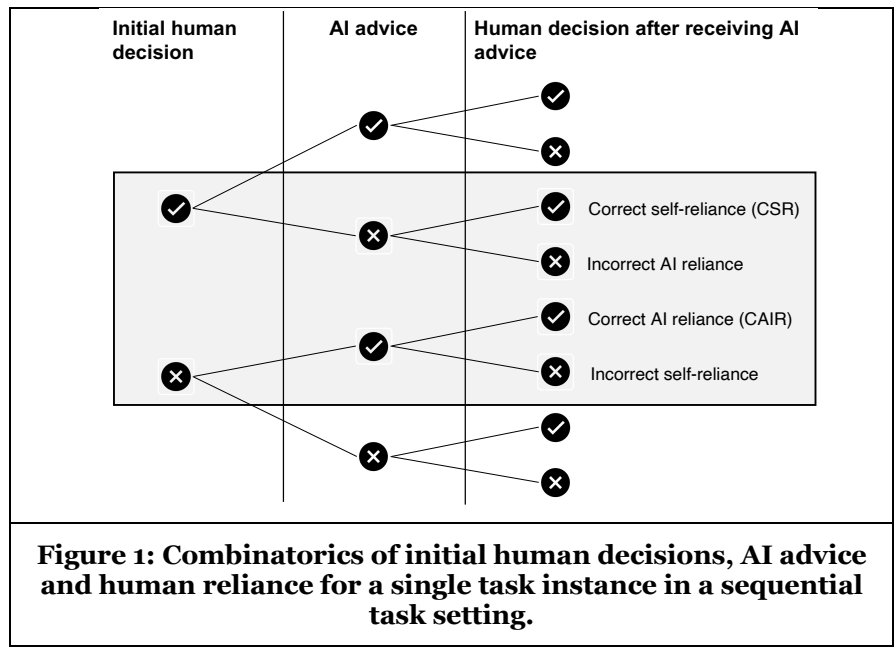

**Figure 1: Combinatorics of initial human decisions, AI advice and human reliance for a single task instance in a sequential task setting.**

Several studies have explored the impact of different explanation techniques on the mental model (Kenny et al. 2021) and AoR, including feature-based (Ribeiro and Guestrin 2016), example-based (van der Waa et al. 2021) and rule-based (Ribeiro and Guestrin 2016) explanations. Some empirical evidence has shown that XAI explanations can help humans differentiate better between correct and incorrect AI predictions (Buçinca et al. 2020), but they can also be misleading, convincing humans to follow incorrect AI advice and leading to poorer team performance (Bansal et al. 2021; Poursabzi-Sangdeh et al. 2018). This ambiguous prospect highlights the importance of further research on AoR and its underlying factors.

Nourani et al. (2020) show that the effect of XAI explanations on humans' reliance is dependent on their initial domain knowledge. In this work, we explore the potential impact of learning from AI as a mediator between explanations and AoR.

### *Learning from Artificial Intelligence*

In recent years, researchers have been exploring the potential for AI to augment human learning (Cakmak and Lopes 2012; Edwards et al. 2018). One field of research that has emerged is machine teaching (Zhu et al. 2018). In this field, instead of assisting humans in decision-making, AI systems are set up as learning systems to teach humans. One example are AI systems that train crowd-sourcing workers to correctly annotate images (Wang and Vasconcelos 2020). The concept involves selecting the optimal teaching set (Zhu et al. 2018) to achieve the best learning performance (Singla et al. 2014).

A central focus of research in this field is identifying the types of knowledge that can feasibly be captured and preserved. Researchers differentiate between explicit and tacit knowledge. Many scholars underscore that explicit knowledge is articulable and can be conveyed through language (Nonaka and Takeuchi, 2007). According to Alavi and Leidner (2001), this type of knowledge can be documented, such as in a product manual. Conversely, tacit knowledge is more elusive, representing the practical skills or intuition that individuals possess for executing specific tasks (Ryle, 1945). This form of knowledge is challenging to articulate and even more difficult to transfer, as it comprises technical skills, experiences, and human intuition that are not easily put into words (Nonaka, 1994). Research has shown the potential of AI-based learning systems to capture explicit as well as tacit knowledge from data (Stein et al. 2013; Kaadoud et al. 2022).

Recent studies have utilized XAI to generate explanations in such learning systems to transfer explicit and tacit knowledge to novices. The AI system provides additional explanations to the human to improve their knowledge in a specific domain (Alipour et al. 2021). For example, Goyal et al. (2019) use a convolutional neural network for various image classification tasks and apply comparative examples to provide visual explanations. The authors select images with minor changes from another class to generate comparative examples.

In general, XAI techniques can be differentiated in terms of their scope, i.e., global or local explanations (Adadi and Berrada 2018): Global XAI techniques deal with holistic explanations of the models as a whole. In contrast, local explanations work based on individual task instances. Local approaches can be based on examples, features, or rules. Example-based explanations provide examples from historical data that are either from the AI-predicted class (normative examples) or from a different class (comparative examples) (Cai et al. 2019). In an image classification task, a normative example would be an image from the AI-predicted class. A comparative example would be the most similar images from a different class. It is important to note that example-based explanations have a link to ground truth, as they have historically validated labels.

With the advent of research to facilitate XAI in learning systems (Wang et al. 2020; Vandenhende et al. 2022), to the best of our knowledge, there are no studies that investigate how explanations affect human learning in human-AI decision-making. In this study, we investigate how XAI affects human learning without explicitly developing a learning system but by observing its use in human-AI decision-making to improve AoR. Dellermann et al. (2019) argue that humans and AI systems can learn from each other when they collaborate. We follow this line of thought and, following, derive a research model on the potential mediating effect of learning on AoR.AoR.

## Theoretical Development

In this work, we postulate that human learning plays a crucial role in AoR and that it can be influenced by providing explanations. In this section, we derive a corresponding research model that establishes the link between explanations and human learning and its mediating role on AoR.

As a dependent variable, we use the previously introduced tuple of AoR, which comprises the two dimensions of relative self-reliance (*RSR*) and relative AI-reliance (*RAIR*). *RSR* encompasses the cases where the human is initially correct, receives wrong advice, and rightly dismisses it. In this case, the humans' complementary knowledge is leveraged by correcting an AI's wrong output on an instance level. In contrast, *RAIR* encompasses the cases where the human is initially incorrect, gets correct advice, and rightly follows. In this case, complementarity potential from an AI can be exploited as the human would not have been able to correctly solve the task instance without the help of the AI advisor.

We now first derive hypotheses related to the potential impact of learning on AoR and thereafter focus on the impact of providing explanations on learning and potential direct effects.

First, we hypothesize the effect of human learning on RSR and RAIR. Distinguishing these two dimensions allows for a deeper understanding of the underlying mechanisms. We base our hypotheses on theories of domain knowledge (Nourani et al. 2020). Much of the latter work in appropriate reliance has examined the differences between experts (high domain knowledge) and lay workers (low domain knowledge) in terms of their reliance behavior in human-AI decision-making. Research has shown that experts have higher explicit and tacit knowledge and can use this to improve their decision-making (Lebovitz et al. 2021). We argue that the differences between experts and lay workers can be seen as an analogy for learning domain knowledge.

RSR essentially refers to the ability to override incorrect AI advice. In other words, if humans can correctly solve a task instance and receive an incorrect AI recommendation, the RSR tells us how well they can reject that incorrect advice. Nourani et al. (2020) have shown that experts are better at correcting AI errors than lay workers. Consequently, learning should increase the effectiveness of validation and thus increase the RSR. Therefore, we hypothesize:

**H1a:** Learning increases relative self-reliance (RSR).

Improving RAIR reveals a more complex process. In this setting, the human does not have enough domain knowledge to solve the task instances independently. Using our analogy, they could be considered "lay workers" in these cases. Becoming an "expert" can have two advantages. First, at the task instance level, learning a new pattern based on the explanations received could allow the decision maker to solve the task independently, which, in turn, would increase RAIR. Second, an increase in domain knowledge over time may help decision-makers build a better judgment of their own expertise and thus recognize and follow correct AI advice. (Fügener et al. 2022). Both mechanisms would essentially lead to an increase in RAIR through learning.

**H1b:** Learning increases relative AI-reliance (RAIR).

Next, we discuss the impact of providing explanations on human task-specific learning. Prior research at the intersection of IS and human-computer interaction has leveraged learning systems that are supported by AI to teach humans in an example-based manner. The objective of learning systems is the user's knowledge extension. In human-AI decision-making, the goal is to improve performance (Hemmer et al. 2021). In learning system research, the effect of explanations generated by XAI on humans' learning performance is an evolving research stream (Alipour et al. 2021; Goyal et al. 2019). However, previous research has only examined how explanations can be utilized in learning systems. To the best of our knowledge, there are no studies investigating how XAI affects learning in human-AI decision-making.

In general, explanations hold the potential to stimulate new ways of thinking, which can lead to the generation of new knowledge (Saeed & Omlin, 2023). Prior research indicates that learning performance among humans can be enhanced in learning systems through example-based learning (Basu & Christensen, 2013; Castro, 2008; Stark, 2004). In this study, we examine example-based explanations as both normative and comparative instances. Normative examples embody instances of the predicted class, while comparative examples illustrate instances from a different class. As Cai et al. (2019) explain, normative examples aim to set a standard for the intended class by displaying training instances from that class, while comparative explanations provide a contrast between the AI prediction and the most similar instances from a different class. Example-based explanations are anticipated to be easily comprehensible and prompt causal thinking, enabling individuals to deduce cause-and-effect relationships. Fahse et al. (2022) suggest that the efficacy of example-based explanations can be attributed to their compatibility with human reasoning processes and the minimal cognitive burden they impose on users. Yang et al. (2020) further argue that these explanations align with people's inductive (i.e., bottom-up logic) and analogical reasoning (i.e., drawing comparisons from one instance to another), which helps users understand why certain objects are deemed similar or dissimilar.

The main difference between the use of XAI in a learning system and in human-AI decision-making may be the level of accuracy. In the learning system, the AI is expected to achieve perfect accuracy, whereas, in human-AI decision contexts, such perfect accuracy may not be achievable. This is because learning system designers can use historical data and select a training set where the AI prediction is known to be correct. In human-AI decision-making, it is unclear whether an AI's advice is right or wrong. AI is inherently imperfect in human-AI decision-making. However, we argue that example-based explanations are actually not entirely dependent on the performance of the AI. Example-based explanations have a clear ground truth because they are drawn from the training data set. Therefore, example-based explanations have the potential to increase knowledge even if the AI is wrong. In addition, in learning systems, the AI is more of a "personal assistant," and in contrast, in human-AI decision-making, the AI is rather a "team member" of the human, and the explanations can be understood as an interaction between team members.

To sum it up, we hypothesize that the benefits of normative and comparative examples of learning that are present in learning systems will also be present during human-AI decision-making. Therefore, we formulate:

**H2:** Example-based explanations have a positive effect on human learning during human-AI decision-making.

In addition to our hypothesized mediation effects, we also follow the line of thought of previous work and consider that explanations might influence AoR by improving the mental model as well. For this reason, we formulate additional direct path hypotheses between explanations and AoR that represent additional effects of explanations beyond learning.[2]

Explanations allow insights into the reasoning and decision-making of AI models. In the case of inaccurate advice, these insights might help the human decision-maker to evaluate the validity of such reasoning by checking for its alignment with the universal axioms of the task. This process might, in turn, enhance their knowledge regarding the underlying AI model (mental model) and thus improve validation capability. As *RSR* is increased by the correction of wrong AI advice, explanations that enhance this kind of knowledge would increase the *RSR*.

**H3a:** Example-based explanations have a positive effect on relative self-reliance (*RSR*).

Additionally, it could also be possible to calibrate the human mental model of the AI in such a way that without learning, it is still possible to detect a good recommendation of an AI. Therefore, we hypothesize:

**H3b:** Example-based explanations have a positive effect on relative AI-reliance (*RAIR*).

Figure 2 summarizes all hypotheses in one integrated research model. Lastly, any improvement in AoR should, in turn, improve team performance and, at a certain level, enable CTP.

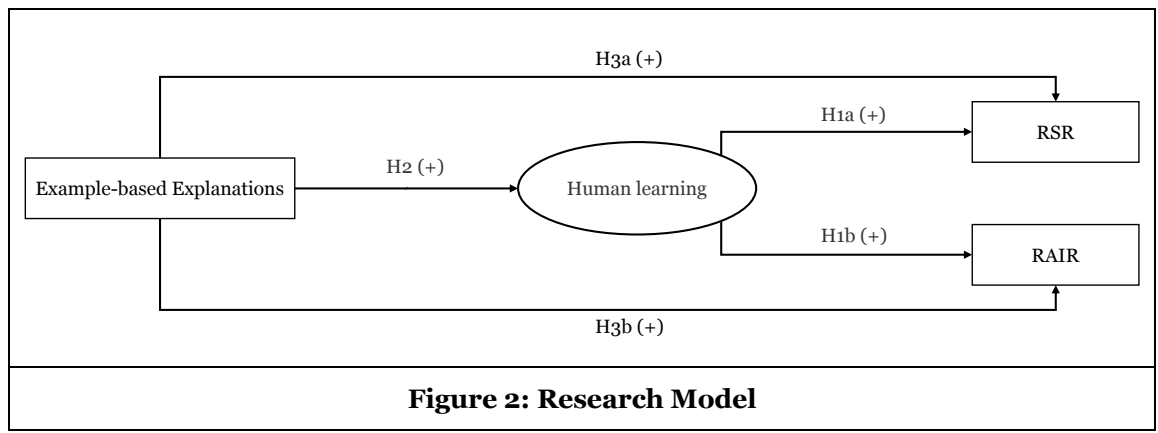

**Figure 2: Research Model**

## Methodology

In this section, we present our design for a behavioral experiment to test our research model.

### Task, Model & Explanations

As an experimental task, we choose a bird species classification task based on image data. We chose the context of image recognition following the reasoning of Fügener et al. (2021): Firstly, image recognition is a broad task that all humans should be capable of executing without requiring specialized skills or training. In behavioral research, the objective is often to establish a context where findings are applicable to various situations. It is believed that observations in general tasks can transfer to more specialized tasks, while contexts that necessitate specific training result in less generalizable outcomes. Secondly, image classification is an area where contemporary AI systems excel (Szegedy et al. 2015), performing at least on par with human abilities (Russakovsky et al. 2015). Thirdly, prior studies have shown the high complementarity potential of image classification (Fügener et al. 2021; Nguyen et al. 2022).

The bird species classification task is based on the Caltech-UCSD Birds-200-2011 dataset (Wah et al. 2011). This dataset has been used extensively in high-profile publications and includes 11,788 images of 200 different bird categories (Goyal et al. 2019; Nguyen et al. 2022. Four black-colored bird classes (American

---

[2] The inclusion of both indirect and direct effect hypotheses in a research model is well known in IS research. For example, see Tereschenko et al. (2022).

Crow, Groove billed Ani, Shiny Cowbird, and Boat tailed Grackle) are chosen for the experiment. In a preliminary study, we tried different combinations of bird classes and the number of classes to get a solvable but still difficult task. Only the consistently black-colored birds are selected from each of these classes, resulting in 216 images in the dataset used. We filter out no-black birds as the preliminary studies have shown that participants tend to choose color as an important factor during learning, which, however, was not an unambiguous feature of the bird classes.

As a model, we use a pre-trained ResNet50 (He et al. 2016) and fine-tune it on our data set. For training, we use Adam as an optimizer with a StepLR rate scheduler with step size 5 and gamma of 0.1. The training data is augmented with random rotation, change in sharpness, and contrast. The model is trained for 9 epochs and achieves an accuracy of 87.96%.

To sample the normative and comparative examples, we follow the approach of (Cai et al. 2019). The comparative examples are searched within the second most probable classes according to the model. Next, we calculate the cosine similarity of the feature vectors of the flattened layer of the model between the target image and all images of the searched class (Chen 2020). Then, the two most similar images are selected. The normative examples are randomly selected from the class predicted by the model. The images which are to be classified in the experiment were omitted from the examples.

### Experimental Design

The experiment is conducted online with a between-subject design where two different conditions are tested (in the following, these conditions are referred to as *baseline* condition and *XAI* condition). Depending on the condition, either just the AI advice or an additional example-based explanation is provided. The study is approved by the University IRB.

Our experimental design is influenced by the requirements to measure learning and AoR. To measure learning, we use standard methods from the AI-based learning research stream (Spitzer et al. 2022; Wang et al. 2023) and conduct two knowledge tests in the experiment. The difference between the two tests then constitutes as learning (Spitzer et al. 2022). Additionally, to measure AoR, a sequential task setup is necessary as it is necessary to measure an initial human decision (Schemmer et al. 2023). These two requirements shape the design of our experiment.

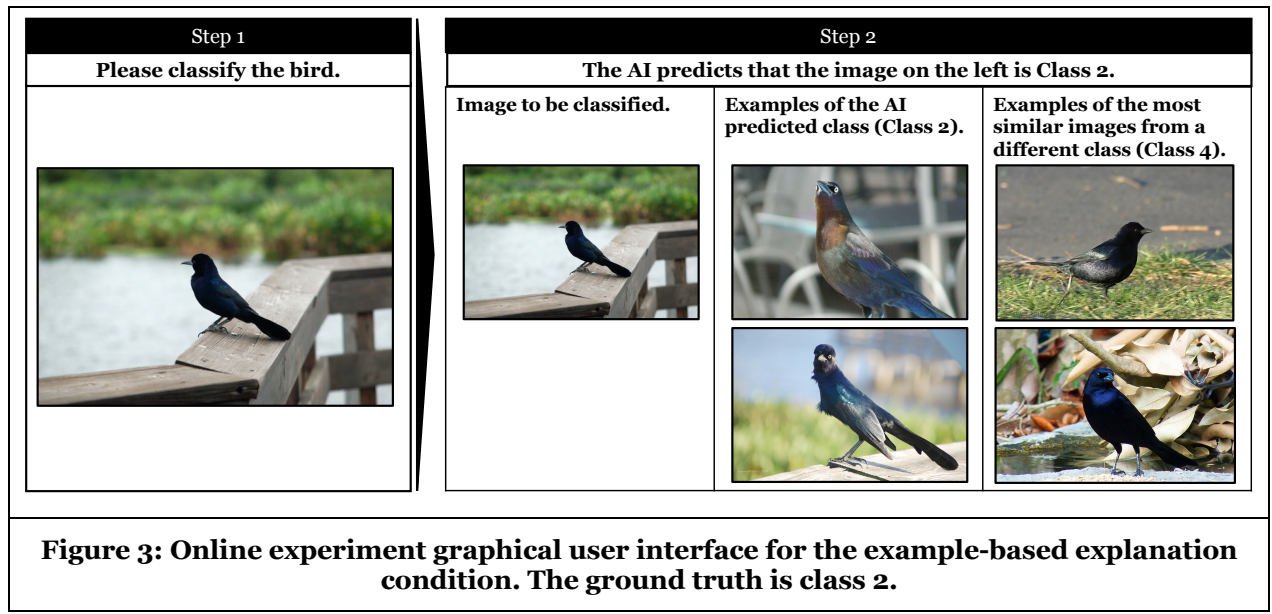

**Figure 3: Online experiment graphical user interface for the example-based explanation condition. The ground truth is class 2.**

Participants are randomly assigned to the condition groups to control for internal validity. The online experiment is initiated with an attention control question. Then, both condition groups receive an introduction to the task. The participants are not provided with any specific performance information about the AI. Then, the participants conduct a tutorial where we show them one example per bird class. After the tutorial, we conduct a knowledge test by asking the participants to classify 8 pictures. Hereby, the 8 pictures are drawn stratified from a sample of 16 images in total.

Once the initial tutorial and first knowledge assessment have been completed, participants move to the main task consisting of 16 individual task instances. We use an advanced sampling strategy to ensure that RSR, as well as RAIR, is possible in our study. We stratified sampled each bird class and control for 50% correct predictions and 50% incorrect predictions. For the AoR measurement concept, sequential task processing is essential. In our study, this means the human first receives an image without any AI advice (see step 1 in Figure 3). Then, the participant is asked to classify the image. Following that, the AI either receives a simple AI advice statement, e.g., "the AI predicts that the image below shows Class 4" or the AI advice and additional example-based explanations (see step 2 in Figure 3). After receiving the AI advice, the participant can change the initial decision. This sequential two-step decision-making allows us to measure AoR. During the main tasks, the participants do not receive feedback on their performance.

After finishing the main task, again, their task-specific knowledge is assessed. Additionally, data on demographic variables are collected.

## *Measurements*

In this work, we measure initial task knowledge by counting the number of correctly classified images in the first knowledge test. Learning is measured as the difference between the number of correctly classified images in the second knowledge test and the first knowledge test.

We measure the AoR following the work of Schemmer et al. (2023) as a tuple of RSR and RAIR. RSR is hereby measured as the number of cases of correct self-reliance divided by the total number of cases in which a previously correct decision-maker receives incorrect AI advice. Correct self-reliance (CSR) hereby is "1" if, in this particular instance $i$, the initial human decision was correct, the AI advice was incorrect, and the human decision after receiving AI advice is correct. Incorrect advice (IA) is "1" if the initial human decision for a task instance $i$ was correct and the AI advice was incorrect.

$$Relative\ self-reliance\ (RSR) = \frac{\sum_{i=0}^{N} CSR_i}{\sum_{i=0}^{N} IA_i}$$

RAIR is the ratio of the number of cases in which humans rely on correct AI advice, and the decision was initially not correct, i.e., in which humans rightfully change their minds to follow the correct advice. Correct AI reliance (CAIR) hereby is "1" if, in this particular case $i$, the original human decision was wrong, the AI recommendation was correct, and the human decision after receiving the AI recommendation is correct, and "0" otherwise. Correct advice (CA) is "1" if the original human decision is wrong, and the AI advice is correct, regardless of the final human decision, and "0" otherwise.

$$Relative\ AI-reliance\ (RAIR) = \frac{\sum_{i=0}^{N} CAIR_i}{\sum_{i=0}^{N} CA_i}$$

## *Participants*

The experiment was performed in April 2023 on the platform "Prolific". Image classification is often done as crowd work and previous IS research has shown the validity of using online studies for image classification (Fügener et al. 2021).

Overall, 100 participants were recruited, 50 per condition. All of them passed our attention check. To incentivize the participants, they were informed that for every correct decision, they get an additional 5 Pennies in addition to a base payment of 1.5 Pounds for an estimated study time of 15 min. Average duration of the experiment was 9:00 minutes in the baseline condition and 12:36 minutes in the XAI condition. Roughly 64% of the final sample identified as being female, almost 36% identified as being male, and one participant preferred not to report. Participants' age ranges from 18 years to 76 years, with an average age of approximately 32 years.

We excluded one participant in the baseline condition because of a failed manipulation check. In addition, some participants had no cases where the AI was correct, and they were previously wrong, leading to an undefined RAIR. We assigned those participants the average value of the condition. Next, after removing missing values, we remove outliers from our data using the z-score method (2 sigma). We find one outlier in the XAI condition with a learning of -5. Additionally, we find 4 outliers with zero correct classifications in the first knowledge test. Which leaves us with a final data set of 48 participants in the baseline condition and 46 participants in the XAI condition.

# Results

In this section we report the results of our behavioral study. First, we provide an overview of the descriptive findings, followed by the results in terms of AoR. We then examine our comprehensive research model, which includes mediations, using structural equation modelling (SEM) and conduct an exploratory subgroup analysis.

### *Descriptive Results & Appropriateness of Reliance*

The descriptive results of our study are presented in Table 1. They are divided according to the experimental condition. We evaluated the significance of the results using t-tests after controlling for normality. Otherwise, we use Mann-Whitney U tests. In order to control for multiple comparisons and reduce the likelihood of Type I errors, we applied Bonferroni corrections to our statistical analyses. First, we report descriptive measures of initial knowledge and learning. Next, we report correlation measures and then our analysis of AoR.

| Treatment | Learning ** (SD) | RSR (SD) | RAIR (SD) |
|---|---|---|---|
| **Baseline** | -0.19 (1.89) | 64.89% (32pp) | 57.82% (39pp) |
| **Example-based explanations** | 0.63 (1.9) | 74.61% (25pp) | 66% (33pp) |
| **Table 1: Descriptive Results (*** p < 0.01, ** p < 0.05, * p < 0.1)** | | | |

Our control variable, initial task knowledge, is not significantly different between groups, meaning that participants start the experiment with the same knowledge on average (baseline mean = 4.54, XAI mean = 4.26; two-tailed t-test: T = - 0.8, p = 0.43). On average, participants correctly classify about 4 of the 8 birds in the initial knowledge test. Note that this does not mean that they perform randomly on average, as we perform a multiclass classification with 4 classes, i.e., random guessing would lead, on average, to a performance of 2 out of 8. Testing against randomness shows that participants in both groups have significant initial knowledge (one-sample t-test: baseline: T = 9.55, p < 0.01; XAI: T = 9.86, p < 0.01). In summary, initial knowledge is not significantly different between the two groups, and participants have sufficient initial knowledge to avoid guessing at random.

Learning is significantly different between the groups (two-tailed t-test: T = 2.09, p = 0.04), which means that, on average people, learned more in the XAI condition. In addition, we test with a one-sample t-test whether the learning is significantly different from 0. In the baseline condition, we find no significant difference (T = -0.69, p = 0.49). In the XAI condition, we observe that learning is significantly greater than zero (T = 2.25, p = 0.03). This means that our example-based explanations not only increase learning but also bring it to a significant higher level than 0.

| Treatment | Correlation | P-Value |
|---|---|---|
| **Learning – RSR*** | 0.18 | 0.07 |
| **Learning – RAIR** | 0.08 | 0.43 |
| **Table 2: Correlation Analysis (*** p < 0.01, ** p < 0.05, * p < 0.1)** | | |

We also analyze the Pearson correlation on paths 2 and 3 of our hypothesis. The results are presented in Table 2. To analyze the correlation paths, we do not yet distinguish between conditions. We find a weak correlation between learning and RSR but no significant correlation between learning and RAIR.

Next, we analyze AoR. Participants' RAIR and RSR are not significantly different between conditions. However, both values are relatively high if we compare them with related literature (Schemmer et al. 2023; Taudien et al. 2022). We also observe that, although not significant, there is a trend that our example-based explanations increase both RSR and RAIR. Figure 4 illustrates our AoR analysis.

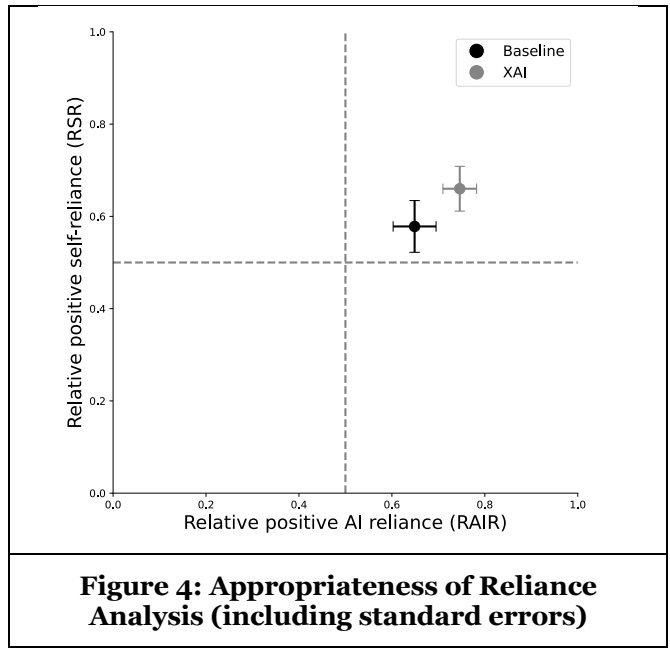

**Figure 4: Appropriateness of Reliance Analysis (including standard errors)**

### Structural Equation Modeling

In addition to analyzing the direct effect of explanations on RAIR and RSR, we use structural equation modeling (SEM) analysis to test our hypothesized research model.

Prior to fitting our SEM, we performed missing data identification, outlier detection, normality testing, and selection of an appropriate estimator. We describe how we identify missing data and remove outliers in the previous section. Shapiro's test for normality indicates that several variables of interest deviate significantly from normal distributions. As a result, we conduct the analysis using an estimator that allows for robust standard errors and scaled test statistics (Kunkel et al. 2019). Thus, we use the MLR estimator (Lai 2018).

Our dependent variables are RAIR and RSR. Since these dependent variables are between 0 and 1, we used a logistic model in the Lavaan package, version 0.6-9, in R (Rosseel, 2012). This model has an excellent overall fit (see Table 3). The results for each independent variable are discussed below and visualized in Figure 5.

| | RMSEA | CFI | TLI | SRMR |
|---|---|---|---|---|
| **Measurement Criteria based on (Hu and Bentler 1999)** | < 0.05 | > 0.96 | > 0.95 | < 0.08 |
| **Value** | 0 | 0.99 | 1 | 4.06 |

**Table 3: Structural equation model fitting index using root mean square error of approximation (RMSEA), Comparative Fit Index (CFI), Tucker-Lewis Index (TLI) and Standardized Root Mean Squared Residual (SRMR)**

First, we find a significant relationship between providing explanations and learning (**H2**). We find no direct effect of providing explanations on RSR or RAIR (**H3a & 3b**). We also find no effect of learning on RAIR (**H1b**). However, we do find a strong significant effect of learning on RSR (**H1a**). This reveals a rare but possible phenomenon where the direct effect of a mediation is not significant, but the indirect pathways are. This may occur if learning does not fully mediate the effect of examples on RSR, and a confounder reduces the overall effect. For example, this confounder could be aversion. Future studies should take this into account. We also test the influence of our control variable initial domain knowledge. We find that initial domain knowledge has a positive significant effect on both learning and RSR. Additional runs including the control variables age and gender did not change the effects.

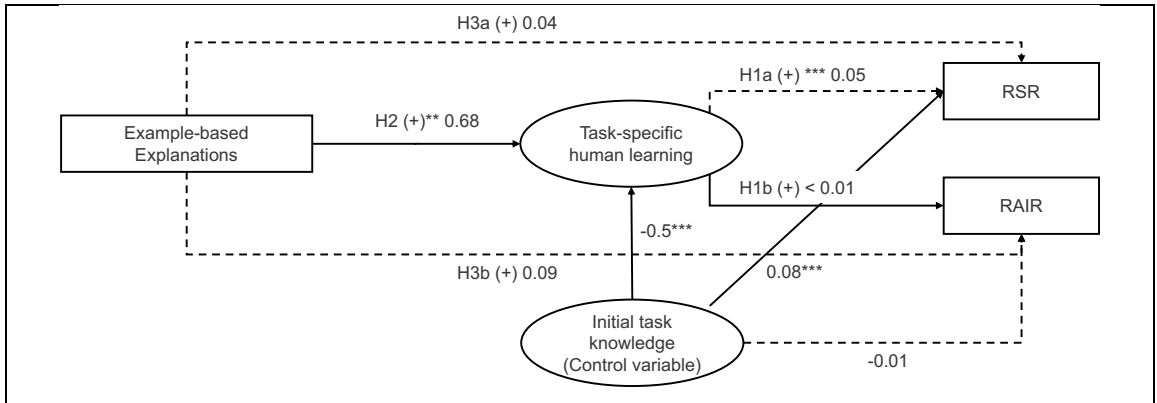

**Figure 5: Structural equation modeling results.**

**Significance: (*** p < 0.01, ** p < 0.05, * p < 0.1)**

### Explorative Sub-Group Analysis

In this section, we want to explore the interesting result that learning does not affect RAIR at all. Therefore, we analyze both conditions (baseline and XAI) separately and perform an exploratory subgroup analysis. Overall, we find significant differences between the SEMs fitted to the different conditions (X2 = 18.26, p = 0.02). Table 4 shows the path coefficients and p-values of the subgroup models.

|  | Baseline Condition | | XAI Condition | |
|---|---|---|---|---|
| **Path** | **Estimate** | **P-Value** | **Estimate** | **P-Value** |
| **Learning – RSR** | 0.08 | < 0.01 | 0.04 | 0.14 |
| **Learning – RAIR** | -0.06 | 0.08 | 0.05 | 0.05 |
| **Table 4: Subgroup Analysis** | | | | |

In the XAI group, we find our expected positive significant effect of learning on RAIR (**H1b**). The analysis of the baseline condition reveals the reasons for the overall non-significant effect. In the baseline condition, the effect of learning on RAIR is also weakly significant, but the path coefficient is negative. This is an interesting result that deserves discussion. In order to answer this question, we need to understand what learning in the baseline condition actually means. Objectively, it is almost impossible to learn anything from the AI in the baseline condition as there is no reference to ground truth. This means that in the baseline condition, there is a tendency for humans to learn nothing or even unlearn patterns that were learned in the tutorial (which is reflected by the overall negative learning value). The limited learning potential during human-AI decision-making could then lead to automation bias (Goddard et al. 2014), i.e., blindly relying on AI advice. Automation bias, in turn, increases RAIR (Schemmer, Hemmer, Kühl, et al. 2022). This could mean that learning in the baseline condition simply means less automation bias and, therefore less RAIR.

To conduct a first validation of this reasoning, we test the impact of learning on absolute reliance[3] in both subgroups. We find that learning significantly decreases reliance in the baseline condition (Path coefficient = -0.03, Z = -2.38, p= 0.02) and has no influence in the XAI condition (Path coefficient < 0.01, Z = -0.05, p= 0.96). Moreover, this reliance positively impacts RAIR in the baseline as well as in the XAI condition. This finding confirms our reasoning above. Future studies need to confirm our findings in a controlled environment.

## Discussion

In this work, we investigated two research questions. First, do explanations improve learning from AI during human-AI decision-making, and second, how does learning from AI influence AoR? To answer both research questions, we derived a research model and conducted a behavioral experiment.

### RQ1: The Effect of Learning on Appropriateness of Reliance

To answer our first research question (How does human learning during human-AI decision-making influence the Appropriateness of Reliance on AI advice?), we analyze the impact of learning on RAIR and RSR.

Applying SEM on our research model does highlight a significant effect of learning on the RSR. However, the direct path between explanations and RSR is not significant. This could mean that some confounders are prevalent that hamper the positive effect of explanations on RSR. One of those confounders could be automation bias (blindly relying on AI advice) (Bansal et al. 2021).

Interestingly, we, however, do not find a positive effect of learning on RAIR. Therefore, we conducted an exploratory sub-group analysis and found significant differences between the effect of learning in the XAI condition and the baseline condition. In the XAI condition, the expected positive effect of learning on RAIR is present. Based on our results, we deduce that a certain level of learning is necessary to observe any impact of learning on RAIR.

### RQ2: The Effect of Explanations on Learning

To answer our second research question (How do explanations of AI influence human learning in human-AI decision-making?), we explore the impact of example-based explanations on human learning. We find a statistically significant difference between a baseline condition and the XAI condition. In the baseline condition, which may be the default configuration in many real-world applications, it may be objectively very difficult to learn anything at all from the AI, which is also confirmed by our experiment. In contrast, we observe statically significant learning in the XAI condition.

To reference our work back to IS research, we want to discuss the link to hybrid intelligence. In their seminal paper on hybrid intelligence, Dellermann et al. (2019) discuss the impact of human-AI decision-making on mutual learning. Humans can teach AI new patterns (often referred to as human-in-the-loop systems (Dellermann et al. 2019) or active learning (Hemmer et al. 2022)). However, the human side of mutual learning has been neglected. We complement IS research with the first work on human learning during human-AI decision-making.

Finally, we will discuss the implications of long-term human-AI decision-making for unique human and AI knowledge. Learning is only possible if the "teacher" has some unique knowledge. The question now is whether humans and AIs can reach an equilibrium where the knowledge of both has aligned such that no further learning is possible. In the case of a static AI (and a human not continuing to learn independently of the AI), this could be the case. However, recent IS research (Kühl et al. 2022) has developed a more dynamic perspective on AI. AI models are frequently updated or being re-trained (the authors call this adaptive AI systems). Similarly, humans usually learn on the job, so they are continuously building unique knowledge. Especially in organizations, this is crucial to prevent knowledge loss and foster the distribution of knowledge (Engbom 2019). Therefore, AI, as well as humans, are in a continuous learning process that creates unique knowledge.

---

[3] We measure reliance as the average number of times the participant followed the AI advice.

### Implications for Theory & Practice

**Theory.** Our research contributes to the literature on organizational learning (Levitt and March 1988) as well as appropriate reliance in human-AI decision-making (Schemmer et al. 2023). Even though learning on the job is a widely recognized approach for organizational learning, research has neglected the potential of in-process learning during human-AI decision-making. So far, learning from AI was always considered as part of a knowledge management tool. We, however, show that also in-process learning is possible and thereby opens up new research potential for the machine teaching domain. With regard to appropriate reliance, learning was so far neglected as an impact factor. With our work, we extend the research model of Schemmer et al. (2023).

**Practice.** Our research implies that practitioners can, with the right design, leverage two benefits at ones from human-AI decision-making—upskilling of the workforce and better performance. With our study, we show the potential to learn from each other in a human-AI decision-making. Thus, these insights can be used to guide not only designers of AI systems but also knowledge managers within organizations to enhance the learning of humans.

### Limitations & Future Work

Despite the contributions of this work, also limitations are present. First of all, our empirical work is limited by the choice of task as well as conducting a single study. However, we believe that image classification is a task with much potential for generalization following the arguments of Fügener et al. (2021). There are many real-world situations where humans need to classify images. Tasks can range from low-stakes tasks, such as product quality inspection, to high-stakes tasks, such as cancer detection.

The generalizability of our experimental findings is also limited due to the choice of explanations. Both normative and comparative examples are a special type of XAI that is directly linked to ground truth. However, example-based explanations are state-of-the-art for learning systems (Goyal et al. 2019). Future work should evaluate different example techniques.

In terms of experiment setup, the sequential task structure required for our measurement approach has certain drawbacks, as it alters the task itself. When humans first complete the task independently before receiving AI assistance, they are already mentally prepared and may respond differently than if they received AI advice immediately. However, this is a known challenge in research and not specific to our study in (Schemmer et al. 2023).

Our results highlight a tension between explanations improving learning but potentially also resulting into automation bias. The positive effects of learning we observe could be diminished by automation bias, resulting in a non-significant net impact on AoR. Future research needs to disentangle the effects.

Most importantly, future research needs to investigate the impact of different design features. Future research should evaluate these potential design features to provide practitioners with a toolkit for the effective use of AI.

## Conclusion

Over the last decade, the emphasis in research on AI adoption and acceptance by humans has facilitated the widespread use of AI in everyday life. The now widespread adoption raises the question of how to harness the potential of human-AI decision-making in the best way possible. The true potential of human-AI decision-making lies in leveraging their complementary capabilities in a way that jointly a performance is reached that is superior to individual AI or human performance. Therefore, to pave the way toward effective human-AI decision-making, it is now crucial to focus on appropriateness of reliance that realizes this potential. In this work, we study the effect of learning on appropriateness of reliance in a behavioral experiment. We find first evidence for learning as an impact factor of appropriateness of reliance and show that human learning can be influenced through explanations. Thus, this work contributes to the design of effective human-AI decision-making.